\def\year{2019}\relax
\documentclass[letterpaper]{article} 
\usepackage{aaai19}  
\usepackage{times}  
\usepackage{helvet}  
\usepackage{courier}  
\usepackage{url}  
\usepackage{graphicx}  
\usepackage{multirow}
\usepackage{makecell}
\usepackage{amssymb}
\usepackage{amsfonts}
\usepackage{amsmath}
\usepackage[ruled]{algorithm2e} 
\usepackage{algorithmic}

\begin{document}
%
\title{Label Embedding with Partial Heterogeneous Contexts}
\author{
	Yaxin Shi$^\dag$, 
	Donna Xu$^\dag$, 
	Yuangang Pan$^\dag$,
	Ivor W. Tsang$^\dag$,
	Shirui Pan$^\dag$
	\\ 
	$^\dag$Centre for Artificial Intelligence(CAI), University of Technology Sydney, Australia\\
	\{Yaxin.Shi, Donna.Xu, Yuangang.Pan\}@student.uts.edu.au,
	\{Ivor.Tsang, Shirui.Pan\}@uts.edu.au
}

\maketitle
\begin{abstract}
Label embedding plays an important role in many real-world applications. To enhance the label relatedness captured by the embeddings, multiple contexts can be adopted. However, these contexts are heterogeneous and often partially observed in practical tasks, imposing significant challenges to capture the overall relatedness among labels. In this paper, we propose a general Partial Heterogeneous Context Label Embedding (PHCLE) framework to address these challenges. Categorizing heterogeneous contexts into two groups, relational context and descriptive context, we design tailor-made matrix factorization formula to effectively exploit the label relatedness in each context. With a shared embedding principle across heterogeneous contexts, the label relatedness is selectively aligned in a shared space. Due to our elegant formulation, PHCLE overcomes the partial context problem and can nicely incorporate more contexts, which both cannot be tackled with existing multi-context label embedding methods. An effective alternative optimization algorithm is further derived to solve the sparse matrix factorization problem. Experimental results demonstrate that the label embeddings obtained with PHCLE achieve superb performance in image classification task and exhibit good interpretability in the downstream label similarity analysis and image understanding task.
\end{abstract}

\section{Introduction}
Label embedding, providing representations for labels, has been widely used in object classification~\cite{akata2016label}, image retrieval~\cite{siddiquie2011image} and novelty detection~\cite{wah2013attribute} tasks. Context information, such as label hierarchy~\cite{DBLP:conf/cvpr/RohrbachSS11}, class co-occurrence statistics~\cite{DBLP:conf/cvpr/MensinkGS14}, semantic attributes~\cite{DBLP:conf/cvpr/LampertNH09}, tags and text descriptions~\cite{mikolov2013distributed} have all been exploited to learn label embeddings. As these contexts provide label relatedness in different aspects, they are in good complement to each other for overall understanding of the labels. For example, weasels, a mammal of the genus Mustela, are considered to be related to cats as they share similar visual attributes. While from the perspective of label hierarchy, weasels should be much more related to the skunks as they belong to the same animal family. Therefore, \mbox{it is necessary to} leverage multiple contexts to learn the label embeddings so that it can well capture the label relatedness in multiple aspects.
\begin{figure}[t]
\centering
\includegraphics[width=7.6cm]{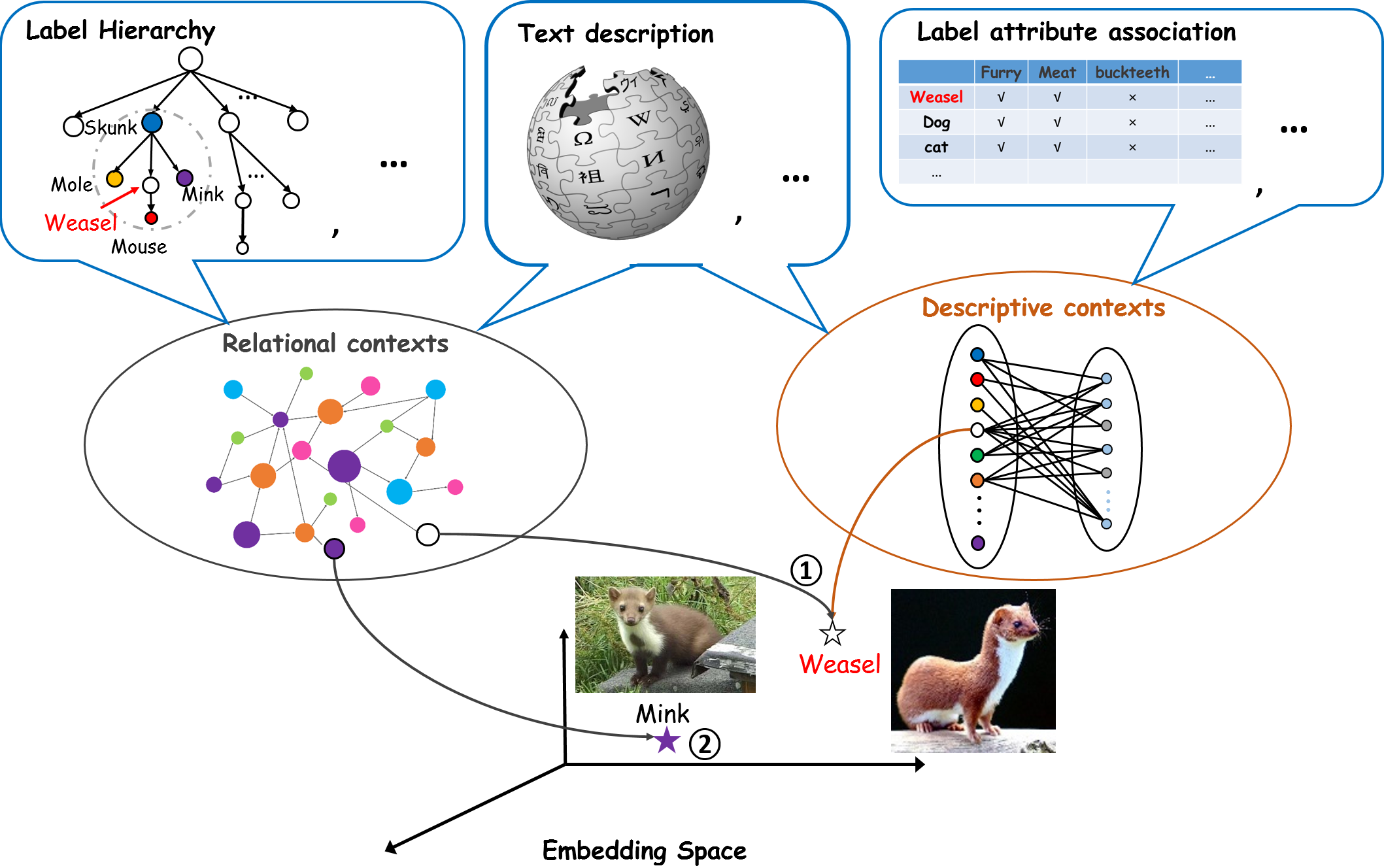}
\caption{\label{fig:motivation} Motivation figure of PHCLE. Stars represent different label embeddings.\textcircled{1} denotes the label embedding for  full contexts label; \textcircled{2} presents that of partial context label.}
\end{figure}

However, learning label embedding from multiple contexts is challenging, due to the heterogeneous nature of the contexts. Generally, based on the relational properties, aforementioned label contexts can be grouped into two basic but heterogeneous categories: relational context and descriptive context. Examples for the heterogeneous contexts is given in Figure~\ref{fig:motivation}. The relational context conveys direct information on label relations. Label hierarchy, class co-occurrence statistics can be grouped into this category. The descriptive context, e.g. attributes and tags, provides the associations between labels and semantic descriptions. Label relatedness is reflected by the sharing descriptions. These two basic categories are universal for existing label contexts. For example, in word2vec~\cite{mikolov2013distributed}, word-context pairs of noun-noun belong to relational context; pairs of noun-adjective and noun-verb belong to descriptive contexts. Universality of the two heterogeneous categories makes it reasonable to formulate multi-context label embedding learning as a general heterogeneous contexts embedding problem. 

To solve the heterogeneous contexts embedding problem, we will have the following challenges:
\begin{itemize}
\vspace{-1mm}
	\item \textit{Challenge 1}: {How to effectively exploit the label relatedness conveyed in each type of the heterogeneous contexts?}
\vspace{-3mm}	
	\item \textit{Challenge 2}: How to align the label relatedness reflected in the heterogeneous contexts?
\vspace{-1mm}	
	\item \textit{Challenge 3}: How can we overcome the partial context problem? As it is difficult to \mbox{obtain label descriptions} \cite{akata2016label}, the heterogeneous contexts are often partially observed in practical label embedding tasks. As shown in Figure~\ref{fig:motivation}, mink lacks the descriptive context.
\vspace{-1mm}	
\end{itemize} 

These challenges cannot be comprehensively solved with existing studies. Former multi-contexts label embedding are mostly obtained via contexts fusion conducted by simple concatenation~\cite{akata2015evaluation} or Canonical Correlation Analysis~\cite{DBLP:conf/eccv/FuHXFG14}. In these approaches, the heterogeneous contexts are treated indiscriminately, so the label relatedness conveyed by the intrinsic property of each context is not fully exploited. Furthermore, as the dependencies between the contexts are not properly captured by their formulation, the contexts are not well aligned. Another representative work for attributed graph embedding, Text-associated Deep Walk (TADW)~\cite{yang2015network}, deals with two heterogeneous contexts within one matrix factorization formula. One severe drawback of TADW is that it requires full correspondence between the contexts, which limits its ability in handling real-world label embedding tasks. In addition, only two contexts can be incorporated in the aforementioned works, which limits their ability to capture more complete relatedness among labels. These two limitations make it desirable to propose a model that can learn label embedding with more partial heterogeneous contexts.

In this paper, we formulate the \textit{multi-context label embedding task} as a general heterogeneous context embedding problem, and propose a novel Partial Heterogeneous Context Label Embedding (PHCLE) approach to solve the challenges. To fully exploit the label relatedness in each context (\textit{Challenge 1}), we tailor-make different matrix factorization formulas to learn the embedding from each type of contexts. To align the labeled relatedness in the heterogeneous contexts (\textit{Challenge 2}), we adopt a shared embedding principle to capture their dependency. A sparsity constraint is further imposed on descriptive context embeddings to select the most discriminative descriptions for better contexts alignment. Due to the adopted shared embedding principle, the proposed PHCLE can handle partial context problem (\textit{Challenge 3}) with an indicator matrix to indicate the missing entries. Furthermore, due to the additive property of the matrix factorization, the proposed PHCLE can be easily generalized to incorporate more contexts. 

Contributions of this work can be summarized as follows:
\begin{itemize}
\vspace{-1mm}
\item We propose a new framework for label embedding, which captures label contexts from two aspects– relational context and descriptive context. Our proposed model is flexible to be generalized to incorporate other label contexts.
\vspace{-3mm}
\item We study the new problem of partial correspondence in heterogeneous context embedding and present a PHCLE model as a solution. Our model captures the label relatedness of each context and aligns them in a shared embedding space via a joint matrix factorization framework, based on which an alternative optimization approach is derived to solve the problem effectively.
\vspace{-2mm}
\item Label embedding obtained with PHCLE achieves superior performance in image classification. Furthermore, the superb interpretability of the obtained label embedding makes PHCLE promising in image understanding tasks.
\end{itemize}
\section{Related Work}
\subsubsection{Label contexts} Context information, such as label hierarchy, class co-occurrence statistics, semantic descriptions and text descriptions, have been widely adopted to learn the label embeddings. Label hierarchy, such as WordNet~\cite{miller1990introduction}, defines the intrinsic structure of labels. Class co-occurrence statistics~\cite{DBLP:conf/cvpr/MensinkGS14} reflects label relations based on the label occurrence rate. Semantic descriptions, i.e. attributes~\cite{DBLP:conf/cvpr/LampertNH09,akata2016label}, provide descriptive information for labels. Relatedness between labels is implied by their common characteristics in the embeddings. Semantic text representations, such as Word2Vec and GloVe, preserve the semantic relatedness between labels based on the text information. As these contexts provide label relatedness in different aspects, it is promising to leverage multiple contexts to learn the label embeddings to capture the overall relatedness among labels. However, as those label contexts are heterogeneous, it is challenging to align multiple heterogeneous contexts in label embedding learning.
\subsubsection{Multi-context label embedding methods} Multiple contexts have been adopted in former embedding works. However, heterogeneity of adopted contexts is not considered. In~\cite{akata2015evaluation}, multiple label embeddings are fused through simple concatenation (CNC). As the embeddings for each context is independently learned, dependencies among the multiple contexts are not captured in CNC. {Canonical Correlation Analysis (CCA) or its nonlinear variants~\cite{DBLP:conf/icml/AndrewABL13}} can be adopted to fuse multiple contexts with the consideration of dependency~\cite{DBLP:conf/eccv/FuHXFG14}. However, due to the intrinsic property of CCA, only the principal component variances of each context is preserved in the common latent space, while context information orthogonal to the principal directions are all lost. Consequently, the relative position of these labels are not well preserved in the obtained label embeddings. For representative multi-contexts network embeddings works, heterogeneity of the contexts is also overlooked. For example, in~\cite{liao2017attributed}, the embeddings are obtained as early fusion for each context conducted via deep neural network. {In ~\cite{DBLP:conf/acl/TuLLS17}, the incorporated contexts are uniformly modeled with softmax formulation.} Other attributed network embedding approaches such as TADW~\cite{yang2015network} and AANE~\cite{huang2017accelerated}, may also be adapted to learn the heterogeneous label embedding. However, these methods typically require full correspondence between different context. This drawback limits their ability in handling real-world label embedding tasks. 
\section{Partial Heterogeneous Context Label Embedding}
In this section, we present the formulation of the proposed Partial Heterogeneous Context Label Embedding (PHCLE). 
\subsection{Notations} 
Let $V_W$ and $V_C$ be the label vocabulary and the context vocabulary in collection $\mathcal{D}$. And \begin{small}${D} \in \mathbb{R}^{|V_C| \times |V_W|}$\end{small} is the co-occurrence matrix constructed via $\mathcal{D}$. \begin{small}${W} \in \mathbb{R}^{n \times |V_W|}$\end{small} and \begin{small}${C} \in \mathbb{R}^{n \times |V_C|}$\end{small} are the label embedding and the context embedding matrices, where $n$ denotes the dimension of the label embedding. \begin{small}${A} \in \mathbb{R}^{|V_W| \times m}$\end{small} is the label-attribute association matrix, where $m$ is the number of attributes. \begin{small}${U} \in \mathbb{R}^{n \times m}$\end{small} denotes the attribute embedding matrix. \begin{small}$\|.\|_{F}$\end{small} denotes the Frobenius norm of the matrix and $\lambda$ is the harmonic factor to balance the components in the formulation.
\subsection{Formulation}
In this paper, we tackle the challenges in label embedding with partial heterogeneous contexts, with three strategies: 1) tailor-made formula for heterogeneous contexts; 2) align partially observed contexts via shared embedding; 3) enhance alignment via discriminative contexts selection.
\subsubsection{Tailor-made formulas for heterogeneous contexts}
For relational context, label relatedness is directly implied by the relationship among labels. Defining label-context as the label relation, Skip-Gram Negative Sampling (SGNS)~\cite{mikolov2013distributed} is representative work that effectively learns label embeddings from label relations. {To enable the incorporation of multiple relational contexts, we adopt an Explicit Matrix Factorization (EMF)~\cite{li2015word} formulation of SGNS, which is given as}
\begin{small}
\vspace{-1mm}
\begin{equation}\label{eq:EMF}
\begin{split}
\min_{{C},{W}} E&MF({D}, {C}^T{W}) \\
\qquad& = -tr({D}^T{C}^T{W}) + \sum_{w \in V_W}\log(\sum_{{d'}_w \in \mathcal{S}_w}e^{{d'}^T_{w}{C}^T{w}}),
\end{split}
\vspace{-8mm}
\end{equation}
\end{small}
where $W$ and $C$ is the label embedding and the context embedding, respectively. {$w\in V_W$ represents a label, ${d}_w \in \mathbb{R}^{|V_C|}$ denotes the explicit word vector of $w$, which also corresponds to the $w^{th}$ column of the co-occurrence matrix ${D}$. $\mathcal{S}_w$ is the Cartesian product of $|V_C|$ subsets, which represents $w$' candidate set of all possible explicit word vectors. $\mathcal{S}_{w,c} = \{0, 1, ..., Q_{w,c}\}$, where $Q_{w,c}$ is an upper bound of the co-occurrence count for the label $w$ and context $c\in V_C$. $Q_{w,c}$ is set to be
\begin{small}$k \frac{\sum_{i}^{|V_W|}d_{i,c}\sum_{j}^{|V_C|} d_{w,j}}{\sum_{i}^{|V_W|}\sum_{j}^{|V_C|}d_{i,j}} + d_{w,c}$\end{small}, where $k$ is the number of negative context samples for each label.}

Specifically, as the label relatedness is conveyed by the label co-occurrence matrix $D$, multiple relational contexts can be exploited by defining multiple label relations~\cite{levy2014dependency} to construct $D$ for specific label embedding task. Particularly, classic SGNS defines linear label-context relation in a large text corpus to learn label representations. However, this context definition fails to consider the structural nature of the labels, thus influence the label relatedness captured by the label embedding. Therefore, label relations that reveal the intrinsic structure of the labels, e.g. label hierarchy, are more suitable to be exploited to construct the co-occurrence matrix $D$ in PHCLE. Consequently, label relatedness conveyed by relational contexts can be properly exploited with the proposed PHCLE. 

For descriptive context, label relatedness is reflected by sharing descriptions. As the descriptive contexts $A$ encodes the label-description associations, it can be effectively modeled with a traditional matrix factorization formula in Eq.~\eqref{eq:ATT}~\cite{DBLP:journals/computer/KorenBV09}, with the two matrices representing the label embedding $W$ and the description embedding $U$, respectively.\\
\begin{small}
\vspace{-2mm}
\begin{equation}\label{eq:ATT}
\|{A} - {W}^T{U}\|^2_{F},
\vspace{-6mm}
\end{equation}
\end{small}
\subsubsection{Align partially observed contexts via shared embedding} As both the two types of heterogeneous contexts are properly modeled, we further adopt a shared embedding principle 
for the alignment of the heterogeneous contexts, which also works with partial contexts. Specifically, as shown in Eq.~\eqref{eq:HCLE_ori}, we formulate the label embedding $W$ to be shared by the two formulas (Eq.~\eqref{eq:EMF}, Eq.~\eqref{eq:ATT}) to capture the dependencies of the heterogeneous label contexts. The adopted principle also enables the proposed PHCLE to handle partial context problem with the matrix $I$ indicating the missing entries.\\
\begin{small}
\vspace{-2mm}
\begin{equation}\label{eq:HCLE_ori}
\min_{{C},{W},{U}}EMF({D}, {C}^T{W})+ \frac{\lambda_{1}}{2}\|\,{I} \odot ({A} - {W}^T{U})\,\|^2_{F}.
\vspace{-4mm}
\end{equation}
\end{small}
\subsubsection{Enhance alignment via discriminative contexts selection} Furthermore, as label descriptions are either manually defined or learned by classifiers, they are often noisy or redundant~\cite{akata2016label}. To better align the heterogeneous contexts, we impose a sparsity constraint on descriptive contexts embeddings~$U$ to select the most discriminative descriptions for contexts alignment. Consequently, the proposed Partial Heterogeneous Context Label Embedding (PHCLE) model can be formulated as
\begin{small}
\vspace{-2.5mm}
\begin{align}
\min_{{C},{W},{U}}E&MF({D}, 
 {C}^T{W})+ \frac{\lambda_{1}}{2} \|\,{I} \odot ({A} - {W}^T{U})\|^2_{F} + \lambda_{2}\|{U}\|_{1} \nonumber\\
&+\frac{\lambda_{3}}{2}(\|{W}\|^2_{F} + {\|{U}\|^2_{F}}) \label{eq:HCLE},
\vspace{-18mm}
\end{align}
\end{small}
{In Eq.~\eqref{eq:HCLE}, heterogeneous label contexts are jointly modeled within a unified matrix factorization framework. This formulation can be detailed explained as follows. (1). For relational context, the EMF preserves the label proximity with the replicated softmax loss~\cite{hinton2009replicated}; (2). For descriptive context, the matrix factorization preserves the label relations implied by the label-description associations. (3). A shared label embedding variable is introduced to achieve consistency for the joint factorization, which contributes to the alignment of the contexts. (4). An indicator matrix is adopted to indicate the partial contexts in PHCLE. Based on matrix completion theory, the factorization can handle missing values~\cite{DBLP:journals/pami/HuZYLH13}. Consequently, partially observed contexts can still be aligned with the proposed PHCLE framework.} 

Conclusively, PHCLE exploits the label relatedness conveyed in heterogeneous context by tailor-made formulas and achieves contexts alignment via a shard embedding principle which also works in partial contexts setting. In this way, the proposed PHCLE framework simultaneously tackles the three challenges in the multi-context label embedding problem, resulting in label embeddings that well preserves the label relatedness conveyed in the heterogeneous contexts.
\section{Model Comparison and Generalization}
One nice property of PHCLE is that it overcomes the partial context problem which can not be solved by existing methods. Furthermore, it is also flexible to incorporate more contexts into our PHCLE model.
\subsection{Comparison with existing methods}
There are some attributed graph embedding approaches \cite{yang2015network,huang2017accelerated,DBLP:conf/ijcai/PanHLJYZ18,DBLP:conf/ijcai/0001PLOS18} that can be adapted to handle heterogeneous contexts. However these methods cannot handle the partial context problem due to the full context correspondence requirement in their formulation.

\begin{figure}[t]
\centering
\includegraphics[width=6.5cm]{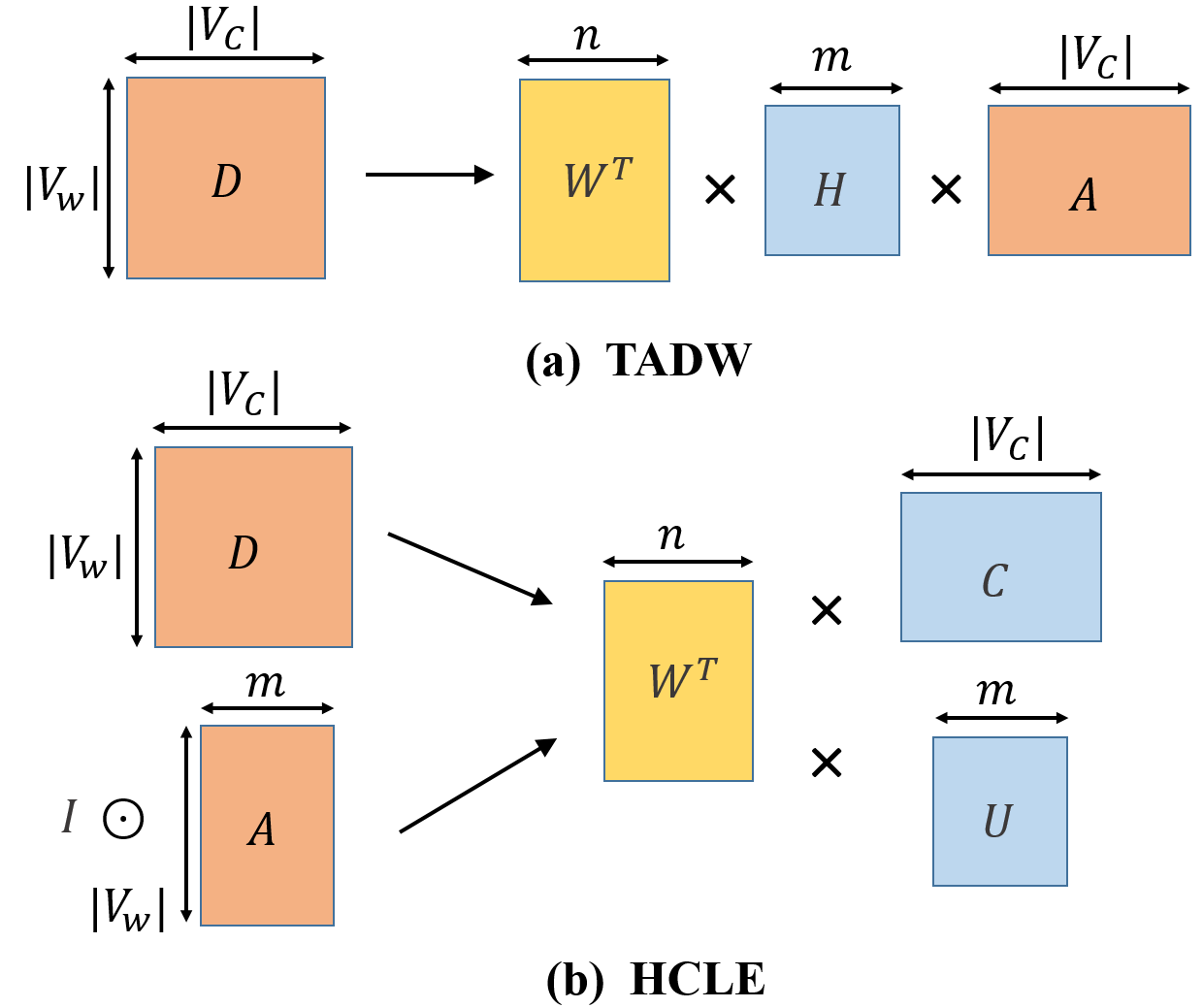}
\caption{\label{fig:matrix} Comparison between TADW and PHCLE. $D$ and $A$ are heterogeneous contexts, $W$ is the learned label embedding, and the others ($H$, $C$ and $U$) are auxiliary matrices.}
\end{figure}

TADW \cite{yang2015network}, for example, learns the embedding $W$ from two heterogeneous contexts $D$ and $A$ within  a single matrix factorization  formula in Eq.~\eqref{eq:TADW}.  
\begin{small}
\vspace{-2mm}
\begin{equation}\label{eq:TADW}
\begin{split}
\min_{{W},{H}} \|{D} - {W}^T{HA}\|^2_{F}
&+\frac{\lambda}{2}(\|{W}\|^2_{F} + {\|{H}\|^2_{F}}),
\end{split}
\vspace{-7mm}
\end{equation}
\end{small}
Obviously, for a label without descriptive contexts, the relational contexts for that label are also dropped. In contrast, in PHCLE, heterogeneous contexts are modeled with a tailor-made formula, with a shared label embedding matrix to capture their dependency (as shown in Figure~\ref{fig:matrix}). Consequently, the two contexts are independent given the label embedding matrix. Thus, our method is more general and flexible to handle partial heterogeneous contexts settings.  
\subsection{ Model generalization}
Due to the additive property of matrix factorization, our PHCLE model can be easily generalized to jointly embed multiple label contexts which belongs to these two heterogeneous categories. Assume there are a total of $n$ relational label contexts, with the constructed co-occurrence matrix denoted as \begin{small}
${D}^{(i)}$\end{small}, and $m$ descriptive label contexts, each denoted as \begin{small}${A}^{(j)}$\end{small}. \begin{small}${C}^{(i)}$\end{small} and \begin{small}${U}^{(j)}$\end{small} denotes the corresponding label context embedding and descriptive contexts embedding, respectively. The Generalized Partial Heterogeneous-Context Label Embedding model (GPHCLE) can be formulated as Eq.~\eqref{eq:general_HCLE}.
\begin{small}
\vspace{-1.5mm}
\begin{align}
\min_{{C}^{(i)},{W},{U}^{(j)}}  &\sum_{i=1}^{n}\alpha^{(i)}EMF({D}^{(i)}, {C}^{(i)T}{W}) \nonumber\\ 
& + \sum_{j=1}^{m}\beta^{(j)}||{A}^{(j)} - {W}^T{U}^{(j)}||^2_{F} + \Omega({W}, {C}^{(i)}, {U}^{(j)}) \nonumber\\ 
s.t. \quad & \alpha^{(i)} \geq 0, \mathbf{\alpha^{T}}\mathbf{1}_{m} =1, \mathbf{\beta^{T}}\mathbf{1}_{m} =1,\label{eq:general_HCLE}
\end{align}
\end{small}
where $\mathbf{\alpha} = [\alpha^{(1)},\alpha^{(2)},...,\alpha^{(i)}]$ controls the weight for relational contexts and $\mathbf{\beta} = [\beta^{(1)},\beta^{(2)},...,\beta^{(j)}]$ is the importance factor for the descriptive contexts, $\Omega({W}, {C}^{(i)}, {U}^{(j)})$ represents the regularizers.
\section{Optimization Algorithm}
Given the above formulation, we propose a gradient descent based alternating minimization algorithm to optimize the proposed PHCLE model. For simplicity, we denote the objective in Eq.~\eqref{eq:HCLE} as\begin{small}
$\mathcal{L}({C},{W},{U})$
\end{small}in the following part.

Specifically, $C$ and $W$ are optimized through SGD. The gradients of \begin{small}$\mathcal{L}(C,W,U)$\end{small} with respect to $C$ and $W$ are 
\begin{small}
\begin{align}
\frac{\partial {L}}{\partial{C}}=  &(\mathbb{E}_{{D}'|C^{T}{W}} {D}^{'}-{D}){W}^{T}, \label{eq:grad_a}\\ %
\frac{\partial {L}}{\partial{W}}=  &{C}(\mathbb{E}_{{D}'|C^{T}{W}} {D}^{'}-{D}) + \lambda_{1}(-{U}{A}^{T}\odot{I}  +{U}{U}^{T}{W}\odot{I})\nonumber \\ 
&+ \lambda_{3}{W},\label{eq:grad_b} %
\end{align}
\end{small}
where \begin{small}$\mathbb{E}_{{D}'|C^{T}{W}} {D}^{'} = {Q}^{T} \cdot\frac{1}{1+exp(-{C}'{W})}$\end{small}~\cite{li2015word}.

The subproblem with respect to $U$ is an Elastic Net~\cite{zou2005regularization} problem defined as
\begin{small}
\begin{equation}\label{eq:elastic}
\min_{ U} \lambda_{1}\|{I} \odot ({A} - {W}^TU)\|^2_{F}+ 
\frac{\lambda_{3}}{2}\|U\|^2_{F} + \lambda_{2}\|U\|_{1},\\ 
\end{equation}
\end{small}
Due to the sparsity constraint on $U$, Eq.~\eqref{eq:elastic} can not be directly optimized through SGD. Instead, we adopt FISTA~\cite{beck2009fast} algorithm to update $U$. Let\begin{small}
$g({{U}}) = \frac{\lambda_{3}}{2}||{U}||^2_{F} + \lambda_{2}||{U}||_{1}$ 
\end{small}and\begin{small}
$f({{U}}) = \lambda_{1}||{I} \odot({A} - {W}^T{U})||^2_{F}$\end{small}, then quadratic approximation of Eq.~\eqref{eq:elastic} at given point ${Z}$ is
\begin{small}
\begin{equation}\label{eq:fista_1}
Q_{\tau}({U},{Z}):= f({Z})+<{U}-{Z},\bigtriangledown f({Z})>+\frac{\tau}{2}||{U}-{Z}||_{F}^{2}+g({{U}}), \nonumber
\end{equation}
\end{small}
which admits a unique minimizer as 
\begin{small}
	\vspace{-1mm}
\begin{equation}\label{eq:fista_2}
p_{\tau}({{Z}}) = \arg \min_{{U}} \{\frac{\tau}{2}|| {U}- {K}||_{F}^{2} \ + g({U})\} \nonumber 
\vspace{-1mm}
\end{equation}
\end{small}
\mbox{where $\tau > 0$ is a constant as stepsize and \begin{small}${K} = {Z} -\frac{1}{\tau}\bigtriangledown f({Z})$\end{small}}.

Details of the alternating minimization of Multi-context Label Embedding are summarized in Algorithm~\ref{alg:1}. FISTA optimization for ${U}$ is shown in Algorithm~\ref{alg:2}. Specifically, $L(f)$ is the Lipschitz constant of the gradient $\bigtriangledown \textit{f}$, and is set as
\begin{small}$L(f) = 2\lambda_{max}({W}_{i}{}^{T}{W}_{i})$\end{small} 
in our algorithm. Stopping condition of the algorithm is set as:\begin{small}$\frac{F({\hat{{U}}}) -F({\hat{{U}}})}{F({\hat{{U}}})} < \epsilon $\end{small}, where $\epsilon$ is a small tolerance value ($\epsilon = 0.0001$ in our experiment). 
\subsubsection{Convergence Analysis}
The proposed PHCLE is formulated as joint matrix factorization. Regarding to each variable, ${C}$, ${W}$ and ${U}$, the sub-optimization problem are all convex. Fixing the other variables, convergence of the alternating optimization for each variable is guaranteed~\cite{li2015zero,wang2017attributed}. Therefore, the objective function will converge to the local minimum accordingly. 
\subsubsection{Optimization for GPHCLE}
For the optimization of the generalized model in Eq.~\eqref{eq:general_HCLE}, if the weights for the contexts in each category are all predefined, GPHCLE can be directly optimized with the proposed algorithm. If $\mathbf{\alpha}$ and $\mathbf{\beta}$ are to be learned, the self-weighted mechanism in~\cite{DBLP:conf/ijcai/NieLL17} can be adopted in each sub-optimization problem in Algorithm~\ref{alg:1}. As it is not the focus, we omit the details here. 
\renewcommand{\algorithmicrequire}{\textbf{Input:}}
\renewcommand{\algorithmicensure}{\textbf{Output:}}\begin{algorithm}[h!]\begin{small}
		\caption{\label{alg:1}  Alternating Minimization for PHCLE}
		\begin{algorithmic}[1]
			\REQUIRE 
			co-occurrence matrix~${D}$, label-attribute association matrix~${A}$,  step-size~$\bf{\eta}$,
			Maximum iteration number~${K}$, trade-off factors~$\lambda_{1}$, $\lambda_{2}$, $\lambda_{3}$ and $\lambda_{4}$
			\ENSURE \begin{small}${C}_{K}$, ${W}_{K}$, ${U}_{K}$\end{small}
		\end{algorithmic}
		Initialize $C_0,W_0,V_0$ to matrix of ones; \\
		\While {${i} < {K}$}{
			\textbf{repeat} \begin{small}${C}_{i} = {C}_{i-1} -\eta \frac{\partial L}{\partial{C}}$\end{small} \quad (See Eq.~\eqref{eq:grad_a})\\
			\textbf{repeat} \begin{small}${W}_{i} = {W}_{i-1} -\eta \frac{\partial L}{\partial{W}}$\end{small} \quad (See Eq.~\eqref{eq:grad_b})\\
			Update ${U}$ using FISTA (Algorithm 2) \\ 
			$i= i+1$ \;
	}\end{small}
\end{algorithm}
\renewcommand{\algorithmicrequire}{\textbf{Input:}}
\renewcommand{\algorithmicensure}{\textbf{Output:}}
\begin{algorithm}[t!]\begin{small}
		\caption{\label{alg:2}  FISTA for updating ${U}$ }
		\begin{algorithmic}[2]
			\REQUIRE 
			${A}$: the label-description association matrix;\\
			${I}\odot{W}_{i-1}$: label embedding obtained in last iteration;\\ 
			${U}_{i-1}$: description embedding obtained in last iteration;\\ 
			$L$: the Lipschitz constant of $\bigtriangledown \textit{f}$;\\
			$InnerMaxIter$ : maximum iteration of the algorithm 
			\ENSURE The optimal solution of $\hat{{U}}$
		\end{algorithmic}
		Initialize ${\hat{U}_{0}} = {U}_{i-1}$, ${Z}_{1} = {\hat{U}_{0}}, {t}_{1} = 1$; \\
		\While {${j} < {MaxIter}$}{
			\begin{small}$\hat{{U}}_{j} = {p}_{\tau}({Z}_{j})$\;
				$t_{j+1} =\frac{1+\sqrt{1+4t_{j}^{2}}}{2} $\;
				${Z}_{j+1} = \hat{U}_{j} + (\frac{t_{j}-1}{t_{j+1}})(\hat{{U}}_{j}-\hat{{U}}_{j-1})$\;
				$j= j+1$ \;
				if $\frac{F({\hat{{U}}}) -F({\hat{{U}}})}{F({\hat{{U}}})} < \epsilon $
				\ break;\end{small}
		}
	\end{small}
\end{algorithm}
\section{Experiments}
In this section, we first evaluate the label embeddings with zero-shot image classification task. Then, we demonstrate the label interpretability of PHCLE with two tasks: label similarity analysis and novel image understanding.
\subsection{Experiment setup}
\subsubsection{Setup for PHCLE}\hspace{-2mm} We learn task free label embeddings for the $1000$ labels of ImageNet 2012 dataset~\cite{ILSVRC15}. Incorporated contexts are constructed as follows.

\vspace{0.5mm}
\noindent{\emph{Relational contexts:} We leverage the neighborhood structure characterized by the label hierarchy in WordNet~\cite{pedersen2004wordnet} to construct a co-occurrence matrix $D$ for the $1000$ labels.}

\vspace{0.5mm}
\noindent\emph{Descriptive contexts:} We adopt the attributes given in the two commonly used attributed image classification datasets: Animals with Attributes (AWA) and Attribute Pascal and Yahoo (aPY), as the descriptive contexts. As not all of the 1000 labels have attributes, PHCLE is set with partial contexts.

\vspace{0.2mm}
\noindent{\emph{Parameter setting}: For PHCLE, we set $K=50$, $d=100$, $InnerMaxIter=50$, $stepsize=10^{-5}$. The trade-off parameters are set via ``grid search'' ~\cite{wang2017attributed} from \{$10^{-2}$, $10^{-1}$, 1, $10^{1}$, $10^{2}$\}, for each baseline. The number of negative samples $k$ is set to be 10 for EMF.}
\subsubsection{Baselines} 
\hspace{-2mm} We select three single context label embeddings and four multi-context label embeddings as the baselines: (1) Attribute label embedding (\textbf{ALE}). We use the attribute annotations released with the datasets; (2) Word2Vec Label Embedding (\textbf{WLE}). We \mbox{use the $500$ dimensional word} embedding vector trained on $5.4$ billion words Wikipedia; (3) Hierarchy Label Embedding (\textbf{HLE}). We follow the setting in~\cite{akata2015evaluation}, and construct an $1000$-dimension embedding for each word accordingly; (4) Concatenation (\textbf{CNC}). We use simple concatenation of ALE, WLE and HLE, as in~\cite{akata2015evaluation}; (5) CCA-fused label embedding (\textbf{CCA}). (6). \textbf{TADW}. We obtain TADW label embedding for labels with full context correspondence with the released codes. As TADW requires fully observed contexts, we specifically compare it to PHCLE with full context correspondence (\textbf{PHCLE\_FC}) in the image classification experiment. Furthermore, for label similarity analysis, label embedding obtained by PHCLE without sparsity constraint (\textbf{PHCLE\_NoSp}) is also compared to demonstrate the impact of discriminative contexts selection in PHCLE.
\subsection{Image classification}
We first apply PHCLE to zero-shot image classification task, where the label relatedness is critical for the performance. Specifically, we conduct experiments on the data that overlaps the $1000$ ImageNet labels for the AWA ($26$) and aPY ($22$) datasets. Details of the adopted datasets are presented in Table~\ref{tab:docdis}. We adopt the ResNet features~\cite{xian2017zero} as the image features and apply three representative zero-shot learning methods, ESZSL~\cite{romera2015embarrassingly}, ConSE~\cite{norouzi2013zero} and SJE~\cite{akata2015evaluation}, (all with their default parameters) to all the embedding methods. We adopt the average per-class top-$1$ accuracy~\cite{akata2016label} as the performance metric. 
\begin{table}[h]
	\centering
	\caption{\label{data_description}Statistics for the adopted AWA and aPY datasets.
	}\label{tab:docdis}
	\scalebox{0.78}{
	\begin{tabular}{cccccccc}
		\hline
		\textbf{Dataset} &    $Y$   &  $Y^{tr}$ & $Y^{te}$ &  $Att$ &  $Training$  &    $Test$    &   $Dim$  \\ \hline
		AWA              &   $26$   &   $20$    &   $6$    &   $85$ &     $11090$  &    $4019$    &   $2048$ \\
		aPY              &   $22$   &   $17$    &   $5$    &   $32$ &     $6925$   &    $1333$    &   $2048$ \\ \hline
\end{tabular}}
\end{table}

Table~\ref{tab:resultPC} shows the result of this image classification task. 
PHCLE consistently outperforms baseline embeddings with all three zero-shot learning methods on both datasets. It demonstrates that PHCLE captures better label relatedness compared with other methods. Specifically, CCA performs the worst among all the baselines. It verifies that the label relatedness is not well preserved in the CCA, as the context information apart from the principal directions is lost. In addition, CNC performs comparably to single context label embeddings on AWA dataset, its results on aPY are even worse than other methods. This indicates that the concatenation of multiple contexts may counteract the influence of each of them. The result of CNC on aPY dataset also indicates that context alignment is difficult in this task. The superior performance of PHCLE verifies its robustness in handling heterogeneous context embedding problem.

Table~\ref{tab:resultFC} compares our method (PHCLE\_FC) with TADW in the full context setting. We can see that PHCLE\_FC outperforms TADW on both datasets. In particular, it achieves almost twice the accuracy to TADW on the aPY dataset. It indicates that PHCLE\_FC is superior in heterogeneous contexts alignment with the shared embedding principle. Furthermore, comparing PHCLE and PHCLE\_FC in these two tables, we can see that PHCLE achieves higher accuracy than that of PHCLE\_FC for most of the settings. It clearly indicates that in PHCLE, the labels with partial contexts help to improve the label embedding for labels with full contexts. 
\begin{table}[t]
\begin{center}
  \caption{The classification accuracy obtained with different label embeddings on AWA and aPY dataset.}
  \label{tab:resultPC}
  \scalebox{0.75}{
  \begin{tabular}{c|ccc|ccc}
    \hline
\multirow{3}{*}{\textbf{Methods}} & \multicolumn{6}{c}{\textbf{Datasets}}\\ \cline{2-7} 
                  & \multicolumn{3}{c|}{ \textbf{AWA}} & \multicolumn{3}{c}{\textbf{aPY} } \\ \cline{2-7} 
                  & ESZSL&ConSE&SJE& ESZSL&ConSE&SJE\\  \hline
ALE               & $63.66$& $50.83$&$61.48$   & $52.60$&$\textbf{52.87}$&$49.75$\\
WLE               & $56.98$& $45.92$&$38.63$   & $45.53$&$39.13$ &$38.04$\\
HLE				  & $53.76$& $58.08$&$54.18$   & $49.79$&$43.35$ &$\textbf{52.18}$\\
CNC			      & $63.66$& $57.97$&$74.27$   & $48.71$&$50.78$ &$33.34$\\
CCA               & $47.05$& $35.46$&$47.05$   & $41.94$&$34.34$ &$39.45$\\
\textbf{PHCLE}     & $\textbf{69.11}$& $\textbf{58.43}$&$\textbf{77.47}$   &$\textbf{54.61}$&$\textbf{52.87}$&$50.91$\\ \hline
\end{tabular}}
\end{center}
\end{table}
\begin{table}[t]
\begin{center}
  \caption{\mbox{The comparison of label embeddings with full contexts.}}
  \label{tab:resultFC}
  \scalebox{0.75}{
  \begin{tabular}{c|ccc|ccc}
    \hline
\multirow{3}{*}{\textbf{Methods}} & \multicolumn{6}{c}{\textbf{Datasets}}\\ \cline{2-7} 
                  & \multicolumn{3}{c|}{ \textbf{AWA}} & \multicolumn{3}{c}{\textbf{aPY} } \\ \cline{2-7} 
                  & ESZSL&ConSE&SJE& ESZSL&ConSE&SJE\\ \hline
TADW              & $50.18$& $56.62$&$40.84$   &$28.13$&$25.21$ &$23.48$\\
\textbf{PHCLE\_FC}          &$\textbf{69.08}$&$\textbf{56.77}$&$\textbf{53.33}$   		                             &$\textbf{50.56}$&$\textbf{53.04}$&$\textbf{42.52}$ \\ 
\hline
\end{tabular}}
\end{center}
\end{table}
\subsection{Label similarity and interpretability}
To assess the efficiency of PHCLE in preserving label relatedness,
we analyze the label similarity and interpretability of PHCLE and other baselines embeddings. 
\subsubsection{Label retrieval} 
We first conduct label retrieval task for the labels with partial contexts in PHCLE. Specifically, given each query, we retrieve the top 5 labels regarding to the cosine similarity~\cite{levy2014dependency}. As compared labels are lack of attribute annotations, ALE is not compared here.

As shown in Table~\ref{tab:similarwords_1}, the retrieved labels of PHCLE are all highly relevant to the query \emph{coffeepot}, as they share the functionality to contain drinks. For WLE, ``chiffonier'' and ``washbasin'' can be considered to be relevant to \emph{coffeepot}, as they are all household items. But it is hard to explain the relevance of \emph{fire screen} and \emph{coffeepot} based on human interpretability. The superior performance of PHCLE over WLE indicates that PHCLE better captures the label relatedness owing to the label hierarchy adopted for the relational context. For HLE, \emph{calderon} and \emph{teapot} are also retrieved, but the other three labels are less relevant to \emph{coffeepot} compared with PHCLE. This demonstrates the tailor-made formula performs better to capture label relatedness from relational contexts. {The overall superior results of PHCLE indicate that it successfully captures label relatedness from multiple aspects and even labels with partial contexts benefit from the alignment of two heterogeneous contexts.}
\begin{table}[t]
	\centering
	\caption{\label{tab:similarwords_1} Label retrieval results. The retrieved labels are listed in descending order. Highly relevant labels are highlighted in bold, weakly relevant labels are in normal font, and irrelevant labels are in italics.} 
	\scalebox{0.8}{%
		\begin{tabular}{c|c|c|c}
			\hline
			\textbf{Query label}      & PHCLE & WLE & HLE \\ \hline 
			\multirow{5}{*}{coffeepot} & \textbf{teapot}  &chiffonier  & \textbf{cauldron}  \\
			& \textbf{cauldron}       &  \emph{fire screen}   &  \textbf{ teapot}  \\
			& \textbf{beaker}        &  washbasin &  barrel   \\
			& \textbf{vase}          &  chocolate sauce &  bathtub   \\
			& \textbf{coffee mug}    & \emph{window shade}      &  bucket\\ 
			\hline
	\end{tabular}}
\end{table}
\subsubsection{Clustering visualization}
\begin{figure*}[t]
	\begin{center}
		\centerline{\includegraphics[width = 0.86 \textwidth]{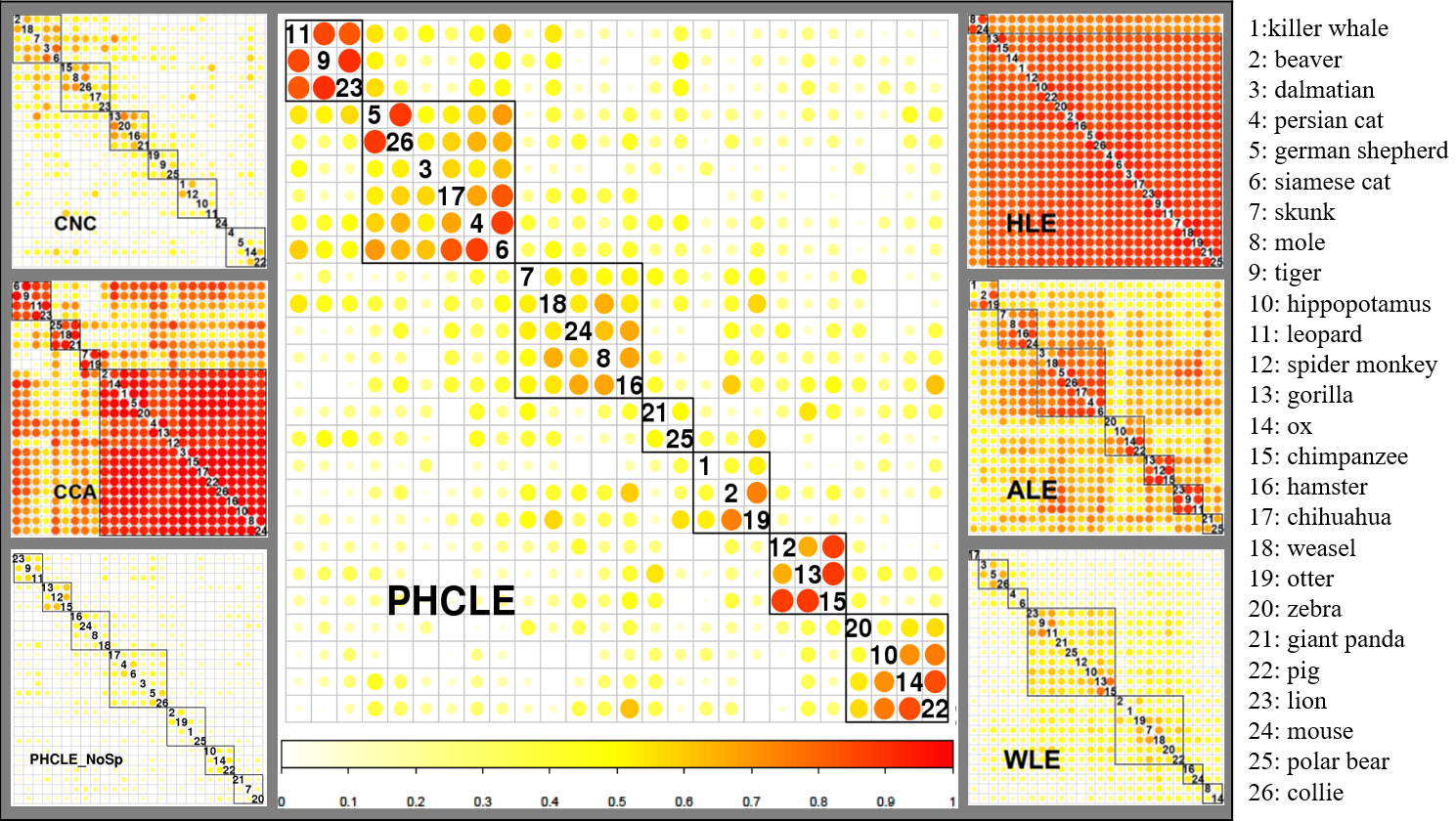}}
	\end{center}
	\caption{\label{fig:conf} Cluster visualization of the correlation matrix constructed via PHCLE and contrastive label embeddings. \textbf{The first column} shows the correlation matrices of multi-context label embedding baselines: CNC and CCA and PHCLE\_NoSp. \textbf{The second column} is the correlation matrix of our PHCLE. \textbf{The third column} presents the correlation matrices of single-context label embedding baselines: HLE, ALE and WLE. Each index corresponds to a label which is listed on the right-hand side.}
\end{figure*}
We further test label similarities for labels with full contexts, based on the cosine similarity. Cluster visualizations for different label embeddings for AWA dataset are shown in Figure~\ref{fig:conf}. The results are quite revealing in several ways.
\begin{itemize}
\vspace{-0.5mm}
\item[1).] For embeddings of single context  (\emph{i.e.}, ALE, HLE and WLE), HLE fails to capture the difference among different embeddings, as the off-diagonal elements show high correlations. The clustering of WLE is not balanced in size, making it difficult to distinguish labels within the big clusters. ALE seems to show superior interpretability among others. However, ALE clusters the \emph{weasel} together with dogs and cats erroneously (in the 3rd cluster), as these animals share the common attributes, such as ``without buckteeth'' and ``eating meat''. Compared with ALE, PHCLE successfully groups \emph{weasel} with its family \emph{skunk} due to its consideration of label hierarchy. Furthermore, many of the off-diagonal elements in ALE are in red color (high correlation coefficients), which indicates its low inter-cluster similarities compared with PHCLE.
\vspace{-0.5mm}
\item[2).]For fused embeddings (CCA, CNC and PHCLE\_NoSp), CCA fails to capture the difference among different labels. For CNC, all the dogs and cats $(5, 26, 3, 17, 4, 6)$ are clustered into three different clusters, with two cats separated. This indicates that multiple contexts aligned with simple concatenation. Our PHCLE produces humanly interpretable clusterings, with the second cluster groups all different species of dogs and cats. For comparison with PHCLE\_NoSp, it is easy to see that PHCLE and PHCLE\_NoSp share similar intrinsic grouping structure, which demonstrates the efficacy of the shared embedding principle for the alignment of the heterogeneous contexts. Furthermore, it is interesting to note that the correlations between the label embedding in PHCLE\_NoSp are very weak, which verifies that the discriminative attribute selection imposed by the sparsity constraint contribute to better alignment of heterogeneous contexts.
\end{itemize}
\subsection{Image understanding}
As PHCLE well aligns the label relations and label descriptions, our PHCLE can be adopted to handle the novel image understanding task. Specifically, for an image which does not belong to any existing classes, we can describe it with relative labels and specific semantic descriptions.  

\begin{figure}[t]
	\begin{center}
		\centerline{\includegraphics[width = 0.5 \textwidth]{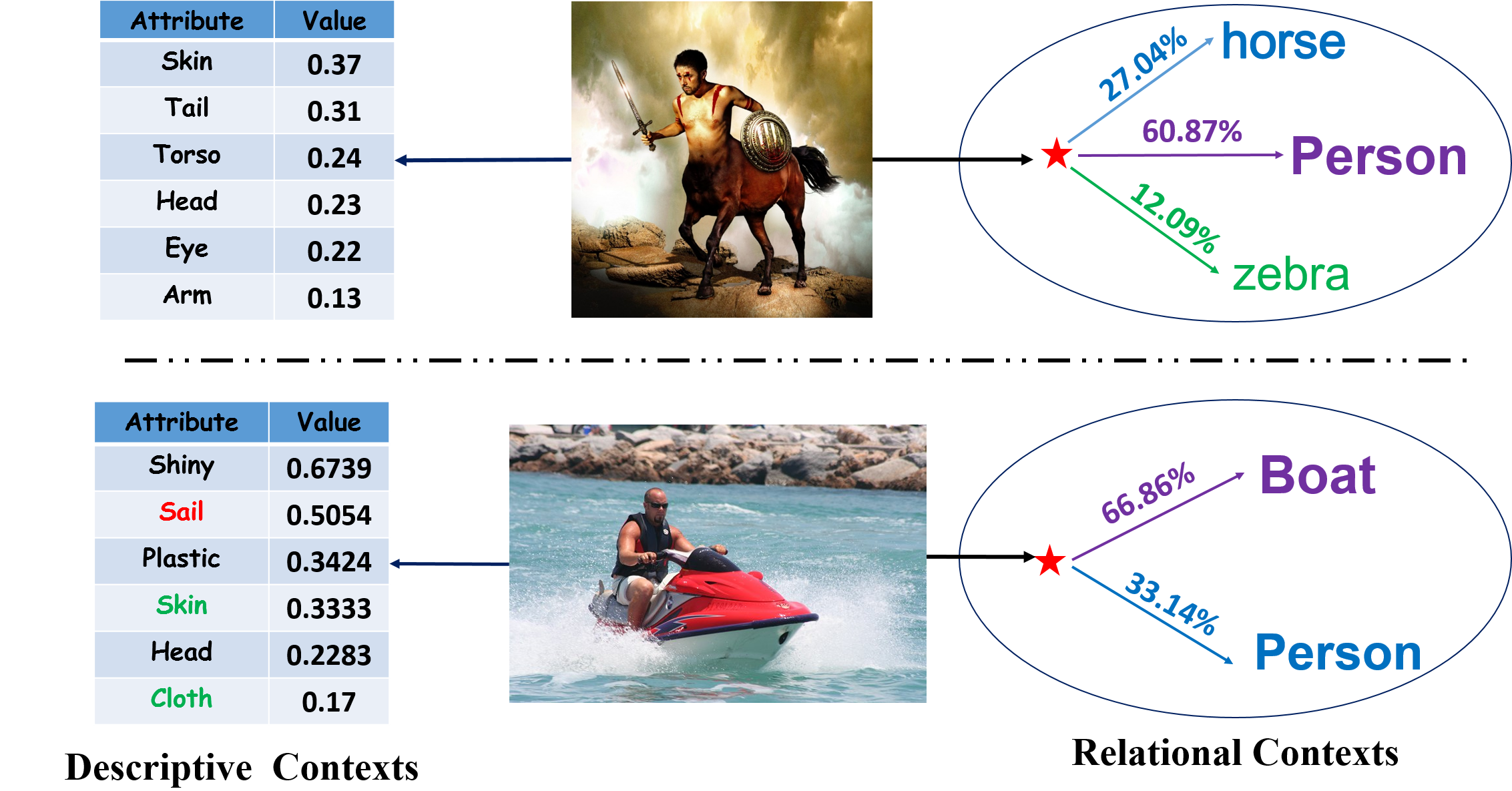}}
	\end{center}
	\caption{\label{fig:attribute}Image understanding with PHCLE. For relational contexts, the digits denotes the images' similarity with its related labels. For descriptive contexts, the attributes in green are creatively exploited, those in red are wrongly predicted.}%
\end{figure}

We conduct experiments on two typical novel classes, centaur and jetski, in aPY dataset. Specifically, we adopt the image-semantic mapping of ESZSL to obtain the semantic embedding of the image $W^{*}$. Then, the related labels are retrieved in the existing label set; the image description ${A}^{*}$ is obtained with $W^{*T}U$. For related labels, top-ranked labels whose cosine similarity account for 80\% of the overall similarity are selected. The obtained similarities are then normalized to get the similarity percentage for each related labels. For attribute description, regarding the predicted value, the top 6 attributes are selected. Figure~\ref{fig:attribute} illustrates the image understanding results for two images. The results are quite revealing for its good human interpretability.
\begin{itemize}
\vspace{-0.5mm}
\item [1). ]For the image of a \emph{centaur} (a mythological creature of half human and half horse), it is described as similar to human and horse, and have specific attributes, such as \emph{skin}, \emph{tail}, \emph{Torso}, et.al. We can observe the result coincides with human's interpretation and the predicted attributes all overlaps the attributes ground truth of the image.
\vspace{-0.5mm}
\item[2). ]
For the image of a person sitting on a Jetski (a recreational watercraft that the rider sits or stands on), it is described to be relative to boat and human. For the semantic description, attributes \emph{Shiny} and \emph{head}, are successfully predicted; \emph{sail} is incorrectly predicted mainly because the image regarded to be most similar to a boat, which also verifies the alignment for heterogeneous contexts in PHCLE. The most interesting is that based on human cognition, \emph{skin} and \emph{cloth} are reasonable to be attributes of the image, but it is not given in the ground truth. This verifies two points: a). human annotated attributes are noisy, making it reasonable to add the sparsity constraint for descriptive contexts selection for better heterogeneous contexts alignment. b). PHCLE achieves good attribute prediction ability due to the alignment of the heterogeneous contexts.
\end{itemize} 
\section{Conclusion and Future Work}
In this paper, we provide a general Partial Heterogeneous Context Label Embedding (PHCLE) framework to solve the three challenges in multiple heterogeneous context label embedding problem. Specifically, we categorize the heterogeneous contexts into two groups, and tailor-make matrix factorization formulas to exploit the label relatedness for each group of contexts. Label relatedness conveyed in those contexts is selectively aligned in a shared space with a shared embedding principle. Due to this formulation, PHCLE can handle partial context problem with an indicator matrix to indicate the missing entries. It can also be easily generalized to incorporate more contexts. Experimental results demonstrate that label embedding obtained with PHCLE achieves superb performance in image classification task and exhibits good human interpretability. 

As descriptive contexts exert huge impacts on relation analysis applications, such as social network analysis and recommendation, we will further study on exploiting label relations conveyed by the descriptive contexts with PHCLE in the future work.
\section{Acknowledgments}
This project is supported by the ARC Future Fellowship FT130100746, ARC LP150100671, DP180100106 and Chinese Scholarship Council (No.201706330075). 
\bibliography{PHCLE}
\bibliographystyle{aaai} 
\end{document}